\documentclass{article}

\PassOptionsToPackage{numbers,sort&compress}{natbib}
\IfFileExists{neurips_2026.sty}{%
  \usepackage[preprint]{neurips_2026}%
}{%
  \usepackage[preprint]{paper/neurips_2026}%
}

\usepackage[T1]{fontenc}
\usepackage[utf8]{inputenc}
\usepackage{microtype}
\usepackage{amsmath,amssymb}
\usepackage{booktabs}
\usepackage{array}
\usepackage{graphicx}
\usepackage{float}
\usepackage[table]{xcolor}
\usepackage[colorlinks=true,linkcolor=blue!50!black,citecolor=blue!50!black,urlcolor=blue!50!black]{hyperref}

\definecolor{tablehead}{HTML}{E8EEF7}
\definecolor{tablesubhead}{HTML}{F3F6FA}
\definecolor{impactgood}{HTML}{E6F4EA}
\definecolor{impactwarn}{HTML}{FFF3D6}
\definecolor{impactbad}{HTML}{FCE8E6}
\definecolor{impactneutral}{HTML}{F2F2F2}

\newcommand{\rpas}{\mathrm{RPAS}}
\newcommand{\ctr}[1]{\mathrm{CTR}(#1)}
\newcommand{\se}[1]{\ensuremath{\,{\scriptstyle\pm\,#1}}}
\newcommand{\paperfigure}[1]{%
  \IfFileExists{figures/#1}{%
    \includegraphics[width=\textwidth]{figures/#1}%
  }{%
    \includegraphics[width=\textwidth]{paper/figures/#1}%
  }%
}

\title{Does Your Agent's Memory Survive a Model Upgrade?\\
A Controlled Study of Memory Portability}

\author{%
  Ankit Goyal\\
  \texttt{angoyal@linkedin.com}
  \And
  Jaideep Ray\\
  \texttt{jaray@linkedin.com}
}

\begin{document}
\maketitle

\begin{abstract}
Model upgrades are routine; memory migrations are not. An agent can keep the same memory store and still forget: a new model may interpret old notes differently, mixed embedding versions may break retrieval, and repair may fail without the original evidence. We compare memory as the same history is preserved verbatim for long-context reading (LC-RAW), divided into chunks for retrieval-augmented generation (RAG), compressed by a model into natural-language notes (NOTES), or normalized into a fixed-schema knowledge graph (KG-fixed). The study uses 48 synthetic histories with randomized answer codes, exact scoring, and two open-weight models with sub 10 billion parameters.

Our measurements show that fixed-schema structures transfer reliably, with KG-fixed accuracy changing by only $+0.0004 \pm 0.0020$ following a writer swap. Conversely, compressed NOTES exhibit high model coupling, with accuracy shifting asymmetrically by $+9.91$ or $-13.28$ percentage points depending on the specific migration direction. In RAG systems, partial embedding migrations using a 50/50 mixed index capture only a 4.96-point accuracy improvement, forfeiting the majority of the 11.90-point gain achieved through full re-embedding. Diagnostic decomposition attributes 80\% ($0.467 \pm 0.014$) of the NOTES accuracy deficit to information lost during initial construction, whereas retrieval failures drive 81\% ($0.364 \pm 0.012$) of the RAG deficit. Finally, store-only repair of NOTES fails to reach a 90\% performance recovery target in all 48 test cases, whereas retaining the raw source history enables successful recovery in 34 of 48 cases for one tested direction. These findings highlight the necessity of direction-specific migration testing, strict embedding space isolation, and the retention of source histories for memory repair.
\end{abstract}

\section{Introduction}
\label{sec:intro}

Agents need memory because useful work rarely fits inside one interaction. Across domains, this means recalling last week's decisions for a project, preserving repository conventions and earlier tool outcomes for a coding agent, and carrying preferences, commitments, and unfinished tasks across sessions. The model's context window provides temporary working context, but it is not durable state. Replaying an entire history on every turn also becomes slow, expensive, and eventually impossible as the history grows. Agent systems therefore persist selected experience outside the model and bring relevant parts back when needed.

That external store becomes part of the agent's behavior. We study four common formats. \textbf{Long-context raw history (LC-RAW)} keeps the full conversation or event history. \textbf{Retrieval-augmented generation (RAG)} stores transcript chunks and retrieves a small set for each query. \textbf{Model-written consolidated notes (NOTES)} stores compressed natural-language summaries; NOTES is simply our label for this format. \textbf{Fixed-schema knowledge graph (KG-fixed)} stores facts as subject--predicate--object claims under a schema shared by every model. Each format makes a different trade-off: raw history uses more context, retrieval can miss the right record, notes can omit evidence during compression, and a fixed schema can represent only the facts it was designed to hold. Memory is therefore not just a database attached to an agent. It is a boundary that can lose information between past experience and future decisions.

The boundary matters because memory and models have different lifetimes. A useful store may accumulate for months or years, while the models around it change much sooner. Teams adopt new language models for capability, latency, cost, context length, safety, or provider availability. They replace embedding models when retrieval quality improves, route requests across several models, and retire older endpoints. Rebuilding every historical memory after each change can be costly, and the original evidence may no longer exist. In practice, the new model often inherits the old store.

Upgrading an agent should not make it forget. Yet a memory migration is not always a simple copy operation. A new reader may interpret another model's notes differently. A new embedder may place records and queries in an incompatible vector space. A repair process may discover that consolidation discarded the only copy of a critical fact. The deployment still starts, the database still returns results, and nothing raises a migration error. Performance simply drops.

This is a different reliability problem from ordinary benchmark accuracy. Developers need to know whether a memory architecture remains usable when one component changes, which shortcuts fail silently, and what source data must be retained for recovery. A store that performs well on day one can still be a poor long-term design if it is tightly coupled to the model that created it.

Prior work has begun to study memory across models. MemCollab reports that naive cross-agent transfer can degrade performance and distills agent-agnostic memory; a trainable graph-memory study tests robustness when GPT-4o or Gemini-2.5-Pro constructs the graph; and procedural-memory work measures skill transfer across models \citep{chang2026memcollab,xia2025graph,belikova2026procedural}. Production systems also support memory import, export, re-indexing, and vector migration \citep{borro2026memori,pinecone2026docs,weaviate2026docs,anthropic2025memory}. But a basic engineering question remains: how do common memory formats compare under the same migration conditions? We still do not know which formats depend most on the model that created them, what an incomplete embedding migration costs, or whether a damaged store can be repaired after the raw history is gone. Appendix~\ref{app:related} gives a fuller comparison.

\textbf{To answer these practical questions, we designed an experimental framework that isolates memory creation from memory reading.} In our setup, the \textbf{writer} turns an agent's history into persistent memory, while the \textbf{reader} later uses that store to answer queries. After an upgrade, model $A$ remains the writer of the existing memory and the new model $B$ becomes its reader.

We evaluate this handoff by isolating one variable at a time: the writer--reader pair, the embedding model, or the source available for repair. For an $A{\to}B$ migration, we compare $B$ reading $A$'s store with $B$ reading a store that $B$ created. This keeps the reader fixed and separates migration loss from the reader's general ability. Randomized answer codes prevent answers from coming from pretraining, and exact scoring avoids the need for a model judge. Before collecting the main results, we saved four planned tests and their five-point decision threshold in a signed Git tag (Section~\ref{sec:lock}).

\paragraph{Contributions.} We provide: (1) a controlled comparison of four common memory formats; (2) separate measurements for each migration direction; (3) tests of model swaps, mixed embedding indexes, and repair with or without raw history; and (4) a breakdown of loss into memory writing, retrieval, and reading, followed by a practical design playbook.

\section{What does it mean for memory to survive an upgrade?}
\label{sec:setup}

\subsection{Separate the components that change}

An agent memory system has four roles: the \textbf{actor} produces the experience, the \textbf{writer} $W$ saves it as memory, the \textbf{reader} $R$ later uses that memory, and the \textbf{embedder} $E$ supports vector search. An upgrade may change any one of these parts.

We use embeddings for RAG only. During a write, the system chunks the history and stores each chunk's vector with its text. During a read, it embeds the query with the same configuration, ranks stored vectors by similarity, and sends the top-ranked text to the reader. The vectors route memories; the reader does not inspect them. A reader swap needs no re-indexing if $E$ stays fixed. An embedder swap requires full re-embedding because old and new vectors occupy different spaces, even when their dimensions match.

We hold the history $X=(x_1,\dots,x_T)$ fixed and vary the part that writes, indexes, reads, or repairs it. For memory format $s$, $S=\mathrm{Write}_s(X;W,E,\beta_w)$ and answers come from $\mathrm{Read}_s(q;S,R,\beta_r)$. This setup shows whether a failure began while saving, retrieving, or reading the memory.

\subsection{Memory formats and their trade-offs}
\label{sec:substrates}

We evaluate four memory formats actively deployed in practice:

\begin{itemize}
\item \textbf{LC-RAW} keeps the complete transcript. It preserves the most information, but every read must process the full history.
\item \textbf{RAG} divides the transcript into chunks and retrieves a small set for each query. It reduces reading cost, but depends on search quality and requires a new index when the embedding model changes.
\item \textbf{NOTES} prompts a model to compress history into natural-language summaries under a byte budget. It is compact and readable, but the writer decides what to omit and how to express what remains.
\item \textbf{KG-fixed} converts history into subject--predicate--object (S-P-O) claims under a fixed schema. The shared structure supports targeted reads, but it can represent only the facts covered by that schema.
\end{itemize}

\begin{figure}[H]
\centering
\paperfigure{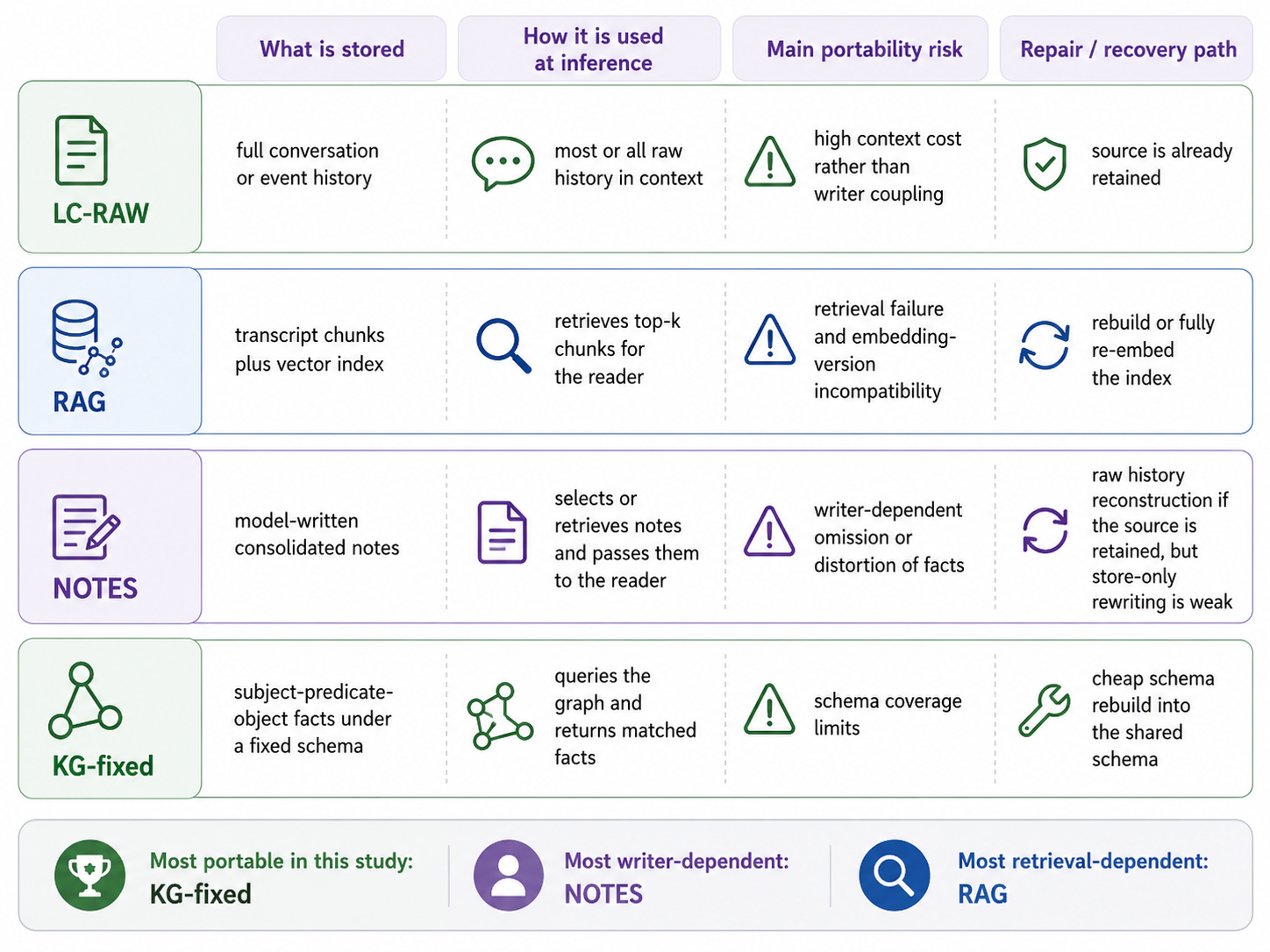}
\caption{Comparison of the four memory formats. Each format preserves different information and introduces a different upgrade risk.}
\label{fig:formats}
\end{figure}

Figure~\ref{fig:formats} compares the persisted state, read path, and primary upgrade risk for each format.

LC-RAW and RAG keep the original source text, so the model cannot change its wording or omit facts while storing it. NOTES and KG-fixed must instead be constructed by a model. In KG-fixed, the schema fixes the record structure, relation names, and source fields. For example, it may require a claim of the form \texttt{project-A -- deadline -- VALUE}, where \texttt{VALUE} is filled with the deadline extracted by the model. In NOTES, the model decides whether to write ``Project A is due Friday,'' merge that fact into another note, shorten it, or leave it out. A reader cannot recover a fact that the writer omitted or that retrieval failed to return.

\subsection{Measure the migration}
\label{sec:metrics}

\paragraph{Retained Performance After Swap (RPAS).} RPAS measures how much performance remains when a reader inherits memory from another model. For a migration from writer $A$ to reader $B$, we first measure $B$ using memory that it created; call this own-store score $N_s(B)$. We then compare it with $B$ using $A$'s store:
\[
\rpas_s(A{\to}B)=\frac{\mathrm{Perf}(s,W{=}A,R{=}B)-\mathrm{chance}}
{N_s(B)-\mathrm{chance}}.
\]
An RPAS of 1 means no performance was lost. An RPAS of 0.8 means the inherited store preserves 80\% of the reader's own-store performance; a value above 1 means the inherited store works better than the reader's own. We also report the direct accuracy gap $\Delta=N_s(B)-\mathrm{Perf}(s,A,B)$. To avoid unstable ratios near zero, we use RPAS only when the own-store score is at least $0.20$ above chance. Importantly, this threshold bound zero times across our primary dataset. Every tested configuration naturally exceeded this performance floor, confirming that our choice of threshold did not artificially filter the reported formats or alter the study's conclusions. We report $A{\to}B$ and $B{\to}A$ separately because the directions can differ.

\paragraph{Cost-to-Recover (CTR).} For threshold $X$, $\ctr{X}$ is the cheapest repair method $\rho$ that restores at least $X\in\{0.90,0.95,0.99\}$ of the new reader's own-store performance:
\[
\mathrm{CTR}_i(X)=\min_{\rho,b}c(\rho,b)\quad\text{s.t.}\quad
\mathrm{Perf}_i(s,\rho_b(S_A),B)\geq XN_{s,i}(B).
\]
Here $b$ is a tested budget rung; $\mathrm{CTR}_i(X)=\infty$ when no method--budget pair reaches the target for history $i$. \textbf{Raw-retained repair} rebuilds memory from the original conversation or event history. The repair model can recover facts that earlier notes omitted or distorted. \textbf{Store-only repair} rewrites the existing store without the source and cannot recreate evidence that is no longer present. Comparing them shows the value of retaining a source copy. We report tokens, graphics processing unit hours (GPU-hours), and dollar cost. Repair tables report both the number of finite values and the median finite cost; the median is therefore conditional on success.

\begin{figure}[H]
\centering
\paperfigure{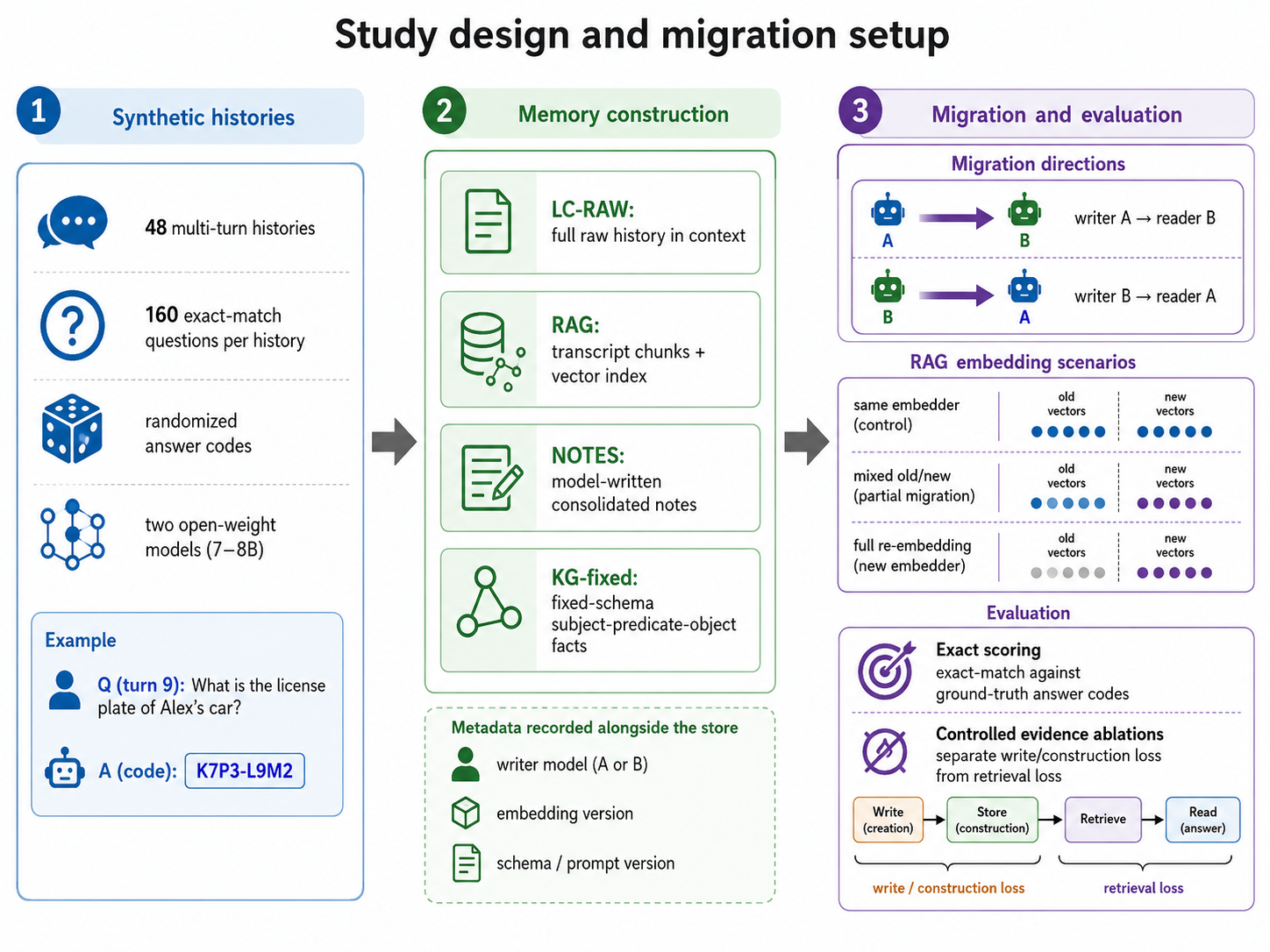}
\caption{Study overview. Fixed synthetic histories are converted into four memory formats. The evaluation changes one component at a time---the model, embedding index, repair source, or diagnostic evidence---and reports exact-match accuracy, RPAS, and CTR.}
\label{fig:overview}
\end{figure}

Together, RPAS and CTR measure what survives an upgrade and what it costs to recover, linking the history-to-store-to-reader pipeline with the model, embedding, and repair changes shown in Figure~\ref{fig:overview}.

\section{Testing memory migrations}
\label{sec:design}

\subsection{Synthetic histories with verifiable answers}
\label{sec:streams}

Real user histories are realistic, but they make it hard to tell why a score changed. Model writing style, pretraining knowledge, subjective labels, and judge behavior can all affect the result. The study instead uses scripted histories with neutral roles and randomized entity and value codes. The agent cannot know the answers from pretraining, and chance accuracy is effectively zero.

Each of the 48 evaluation histories has 160 questions covering direct facts, changes over time, contradictions, multi-step relations, and aliases. Scoring checks the randomized codes exactly; the four planned tests use no large language model judge. The histories are less natural than real conversations, but they let us identify where each fact was lost.

\subsection{Models and upgrade scenarios}
\label{sec:roster}

We pair fixed, locally served versions of two similarly scaled open-weight models: \textbf{Llama-3.1-8B-Instruct} and \textbf{Qwen2.5-7B-Instruct-1M}. This design evaluates a single cross-family migration, not a sample of all model upgrades. Every history fits both context windows with at least 4{,}096 tokens unused. We use deterministic reads and low-temperature writes with fixed seeds. Repeated live reads vary by only 0.0016, far less than the effects reported below.

For the embedding-model migration, we replace \texttt{BAAI/bge-large-en} v1.0 with v1.5. Both models produce 1024-dimensional vectors, so old and new vectors can coexist in one index without causing a dimension error. They still belong to different embedding spaces. The service remains online, but retrieval quality can silently decline.

\subsection{Keep resource limits comparable}
\label{sec:parity}

We fix resource limits wherever the formats permit a direct match. We report every storage quantity in kibibytes (KiB), where $1~\mathrm{KiB}=1{,}024$ bytes, and convert telemetry originally recorded in decimal kilobytes before reporting it. We give both writers the same token allowance, cap NOTES stores at 128~KiB, give every reader the same context budget, and use the same RAG retrieval settings in every comparison. We also match the budget levels for raw-retained and store-only repair.

These controls do not make the memory formats identical. A NOTES reader receives about 111~KiB of consolidated text under the 128~KiB cap, while a KG-fixed reader queries a larger structured store. The NOTES--KG-fixed comparison therefore reflects both the storage format and how the reader accesses it, not the format alone.

\subsection{Validate the evaluation before trusting the result}
\label{sec:calibration}

Before running the main experiments, we validated the evaluation framework to ensure that it behaved predictably. Scores had to fall when memory was removed, recover when the correct evidence was supplied, and decline as the memory was deliberately corrupted. The questions also needed enough score variation to reveal meaningful effects. These checks caught four implementation bugs before the main runs (Appendix~\ref{app:ops}).

Twelve separate histories determined the evaluation sample size. Their variance gave $N=48$ for detecting the planned five-point difference. Calibration also showed that NOTES could not meet the planned 90\% evidence-coverage requirement, so the minimum was lowered to 60\% and the change was recorded. In this check, Qwen's notes retained 85.6\% of the required evidence spans, compared with 64.5\% for Llama. This early gap suggested that the writing stage could strongly affect portability. None of the calibration histories are used in the reported results.

\subsection{Pre-collection fixes and the analysis lock}
\label{sec:lock}

After the early diagnostic work and before collecting the main results, we saved a signed Git tag called \texttt{hypothesis-lock-v1} to fix the study files. Its digital signature lets others check that this version was not changed later.

The saved version listed four planned tests, the expected direction of each result, a five-point threshold, the correction for testing four questions, the use of each history as one independent data point, and the exclusion rules. We retain the original hypothesis labels: Hypothesis 1 (H1), Hypothesis 2 (H2), Hypothesis 4a (H4a), and Hypothesis 7a (H7a). The missing numbers belong to other questions in the wider study plan; they are not missing tests from this paper.

We define the four contrasts explicitly. Let $i\in\{1,\ldots,N\}$ index histories, let $\mathcal{D}=\{\text{Llama}{\to}\text{Qwen},\text{Qwen}{\to}\text{Llama}\}$ be the two migration directions, and let $c$ be the chance-accuracy baseline, which is effectively zero for the randomized answer codes. For memory format $s$, define the per-history retained performance
\[
r_{s,i}(A{\to}B)=
\frac{a^{\mathrm{inherited}}_{s,i}(A{\to}B)-c}
{a^{\mathrm{own}}_{s,i}(B)-c},
\]
and the symmetric writer-swap penalty, in percentage points,
\[
P_{s,i}=\frac{100}{2}\sum_{d\in\mathcal{D}}\left(1-r_{s,i}(d)\right).
\]
A positive $P_{s,i}$ means that inherited memory performs worse than the reader's own store; a negative value means that inherited memory performs better. For the embedding and repair experiments, let $a^{\mathrm{full}}_i(R)$ and $a^{\mathrm{mixed}}_i(R)$ denote accuracy for reader $R$ after full re-embedding and with the 50/50 mixed index, and let $a^{\mathrm{raw}}_i(d,b)$ and $a^{\mathrm{store}}_i(d,b)$ denote NOTES repair accuracy in direction $d$ at matched budget rung $b\in\{1,2,3\}$. The four sample contrasts are
\begin{align*}
C_{\mathrm{H1}}  &= \frac{1}{N}\sum_{i=1}^{N} P_{\mathrm{NOTES},i}, \\
C_{\mathrm{H2}}  &= \frac{100}{2N}\sum_{i=1}^{N}\sum_{R\in\{\mathrm{Llama},\mathrm{Qwen}\}}
\left(a^{\mathrm{full}}_i(R)-a^{\mathrm{mixed}}_i(R)\right), \\
C_{\mathrm{H4a}} &= \frac{1}{N}\sum_{i=1}^{N}\left(P_{\mathrm{NOTES},i}-P_{\mathrm{KG\text{-}fixed},i}\right), \\
C_{\mathrm{H7a}} &= \frac{100}{6N}\sum_{i=1}^{N}\sum_{d\in\mathcal{D}}\sum_{b=1}^{3}
\left(a^{\mathrm{raw}}_i(d,b)-a^{\mathrm{store}}_i(d,b)\right).
\end{align*}
All four estimates are therefore reported in percentage points (pp), and a positive value always favors the planned claim. H1 and H4a are percentage points on the chance-adjusted retained-performance scale; H2 and H7a are accuracy percentage points. Table~\ref{tab:locked} gives the common one-sided test form and the interpretation of each positive contrast.

\begin{table}[h]
\centering\footnotesize
\renewcommand{\arraystretch}{1.12}
\begin{tabular}{@{}llp{6.4cm}l@{}}
\toprule
\rowcolor{tablehead}
Test & Contrast & A positive value means & Run ID \\
\midrule
H1 & $C_{\mathrm{H1}}$ & NOTES loses retained performance after a writer swap. & E1 \\
H2 & $C_{\mathrm{H2}}$ & The mixed index is less accurate than full re-embedding. & E3 \\
H4a & $C_{\mathrm{H4a}}$ & The NOTES writer-swap penalty exceeds the KG-fixed penalty. & E1 \\
H7a & $C_{\mathrm{H7a}}$ & Raw-history repair is more accurate than store-only repair at matched budgets. & E6 \\
\bottomrule
\end{tabular}
\caption{Definitions of the four contrasts recorded in the signed Git tag. The Run ID column preserves the original artifact identifiers rather than the sequential presentation order used in Section~\ref{sec:experiments}. Each test uses $H_0:C_h\leq5~\mathrm{pp}$ against $H_A:C_h>5~\mathrm{pp}$. Repeated directions, readers, and budget rungs are averaged within each history before histories are compared.}
\label{tab:locked}
\end{table}

We fixed only the four tests in Table~\ref{tab:locked} in the tag. Later tests of whether some effects stayed within five points were not part of this planned set.

For each contrast, we apply a one-sample, one-sided $t$ test to the 48 per-history values, testing $H_0:C_h\leq5~\mathrm{pp}$ against $H_A:C_h>5~\mathrm{pp}$. We treat the four raw $p$-values as one family and adjust them with Holm's step-down procedure at family-wise $\alpha=0.05$: after sorting them as $p_{(1)}\leq\cdots\leq p_{(4)}$, we compute $p^{\mathrm{Holm}}_{(j)}=\min\{1,\max_{k\leq j}(5-k)p_{(k)}\}$ and return each value to its original hypothesis. Thus every positive estimate points in the hypothesized direction, but support requires both an estimate above 5~pp and a Holm-adjusted $p<0.05$.

The saved plan called for a bootstrap that repeatedly resamples complete histories and recomputes the same within-history contrasts. We first used the one-sided $t$ tests above instead, then later ran the planned bootstrap with 10{,}000 resamples and the same four-test Holm correction. Both methods gave the same decisions. H2 and H7a passed the five-point threshold; H1 and H4a did not. In the bootstrap, the lower bounds for H2 and H7a were 5.8 and 7.9~pp, and both remained significant after correction ($p\leq4.0\times10^{-4}$).

Because we ran the bootstrap after seeing the $t$-test results, we treat it as a follow-up check rather than the original analysis. Table~\ref{tab:core} reports the first analysis, and Section~\ref{sec:limitations} discusses this limitation.

\subsection{Agent memory migration tests}
\label{sec:experiments}

The evaluation covers four situations that can arise when an agent's model or memory system changes. For readability, we present them as Migration Tests 1--4; the parenthetical run IDs retain the names used by the signed plan and stored artifacts.

\begin{enumerate}
\item \textbf{Migration Test 1 (run E1): A new model reads memory written by the old model.} Each model builds NOTES and KG-fixed stores, and both models read every store. LC-RAW and RAG do not depend on a model writer, so they serve as controls. The test covers both migration directions and includes 576 evaluations across 48 histories.

\item \textbf{Migration Test 2 (run E3): An embedding upgrade partially migrates the index.} Four indexes are compared: the old index, a fully rebuilt index, a 50/50 mixture of old and new vectors, and an ideal router that knows which index contains the answer. The mixed index represents a silent partial migration. The ideal router gives an upper bound and is not a system that could be deployed.

\item \textbf{Migration Test 3 (run E6): A repair process attempts to rebuild the inherited memory.} Under three equal resource limits, the repair test rebuilds NOTES from retained raw history, rewrites NOTES using only the existing store, re-embeds RAG, and rebuilds KG-fixed records into the shared schema. Each repair is compared with fresh memory built for the new model under the same conditions. This test contains 1{,}440 evaluations.

\item \textbf{Migration Test 4 (run E4): Controlled evidence locates where information was lost.} This follow-up test first supplies the correct store item and then the original event. The change in performance helps separate loss during memory construction, retrieval, and reading. The test also examines whether rewriting NOTES in the new model's style helps. This diagnostic test describes the failure stages and was not part of the four planned hypothesis tests.
\end{enumerate}

\section{Results and analysis}
\label{sec:results}

\begin{figure}[H]
\centering
\paperfigure{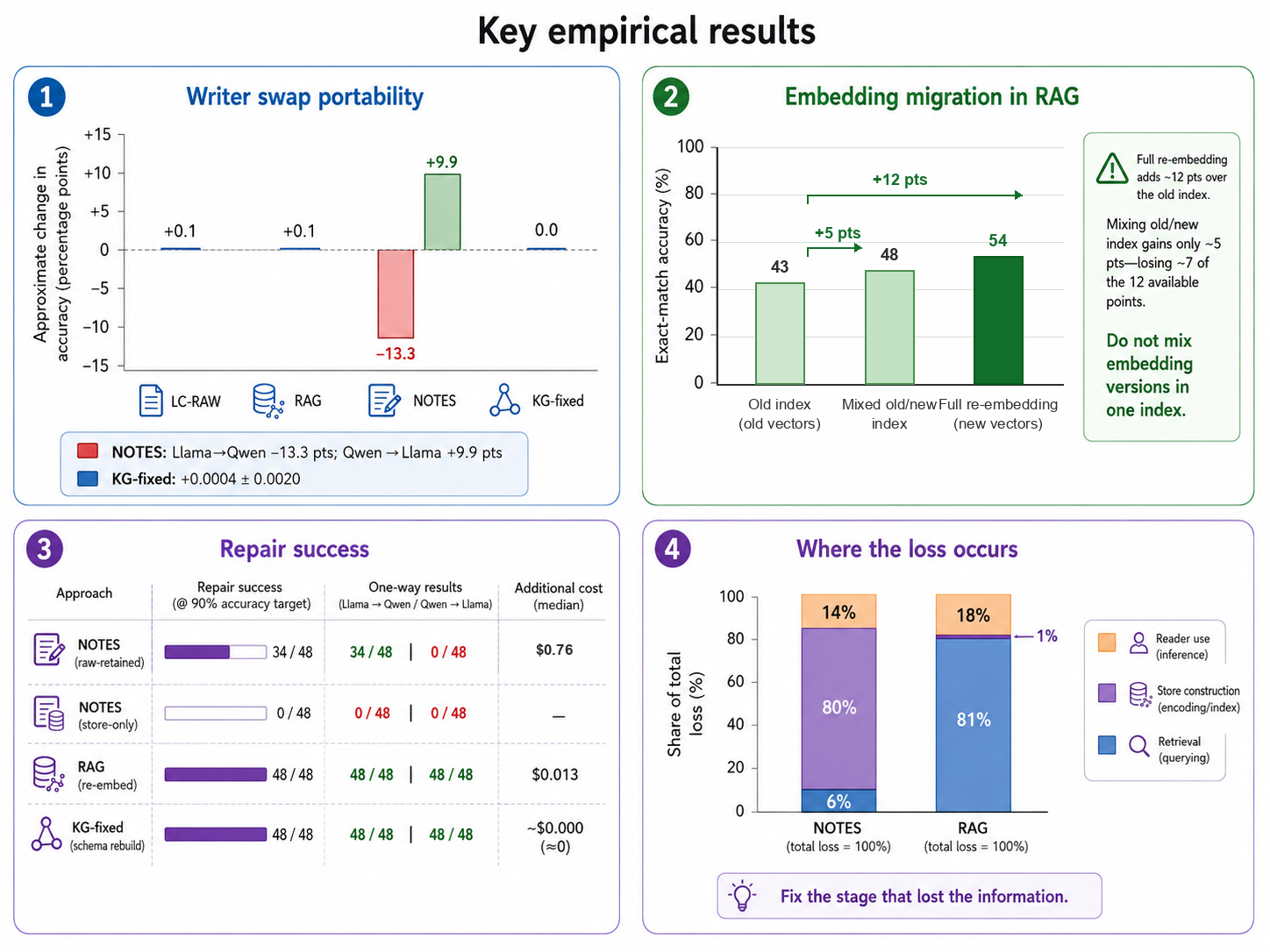}
\caption{Key empirical results. NOTES transfer changes sharply by migration direction, a mixed embedding index leaves seven percentage points unrecovered, NOTES loss is concentrated in memory construction while RAG loss is concentrated in retrieval, and store-only NOTES repair never reaches the 90\% target.}
\label{fig:key-results}
\end{figure}

Figure~\ref{fig:key-results} reports accuracy changes in percentage points (pp), the embedding-migration gap, the stage where information is lost, and repair success.

\subsection{Learnings from the tests}
\label{sec:family}

Table~\ref{tab:core} reports the four tests chosen before the main results were collected using the definitions in Table~\ref{tab:locked}. Two effects exceed the five-point action threshold: using a partly migrated embedding index (H2) and the advantage of raw-history over store-only repair (H7a). The symmetric writer-swap contrasts H1 and H4a do not exceed five points. Their negative estimates mean that the symmetrized retained-performance penalty points slightly opposite to the planned claim; the directional results below show why this average still hides an important failure. Later checks suggest that these averages lie within $\pm5$ points. Unless stated otherwise, $\pm$ is the half-width of a 95\% confidence interval across histories.

\begin{table}[h!]
\centering\scriptsize
\renewcommand{\arraystretch}{1.14}
\begin{tabular}{@{}llrrrl@{}}
\toprule
\rowcolor{tablehead}
Test & One-sided alternative & Estimate & Margin vs. 5 pp & Holm $p$ & Decision \\
\midrule
\rowcolor{impactneutral}
H1 & $C_{\mathrm{H1}}>5$ & $-1.47$ pp & $-6.47$ pp & $1.0$ & Not supported \\
\rowcolor{impactgood}
H2 & $C_{\mathrm{H2}}>5$ & $+6.95$ pp & $+1.95$ pp & $1.3\times10^{-6}$ & \textbf{Supported} \\
\rowcolor{impactneutral}
H4a & $C_{\mathrm{H4a}}>5$ & $-1.51$ pp & $-6.51$ pp & $1.0$ & Not supported \\
\rowcolor{impactgood}
H7a & $C_{\mathrm{H7a}}>5$ & $+8.90$ pp & $+3.90$ pp & $3.9\times10^{-7}$ & \textbf{Supported} \\
\bottomrule
\end{tabular}
\caption{Results for the planned tests across 48 histories. ``Margin vs. 5 pp'' is the estimate minus the action threshold, so positive values clear the threshold. All estimates use the positive-support orientation defined in Table~\ref{tab:locked}. The displayed $p$-values are Holm-adjusted across the four one-sided tests. The later bootstrap gives the same decisions. Later, unplanned checks place H1 and H4a within $\pm5$~pp ($p=0.035$ and $0.037$). H7a exceeds 5~pp but not 10~pp.}
\label{tab:core}
\end{table}

\subsection{A fixed schema transfers better than free-form notes}
\label{sec:kg}

\begin{table}[h!]
\centering\footnotesize
\renewcommand{\arraystretch}{1.14}
\begin{tabular}{@{}llrrr@{}}
\toprule
\rowcolor{tablehead}
Memory format & Reader & Own-store accuracy & Inherited-store accuracy & Swap $\Delta$ (pp) \\
\midrule
NOTES & Llama & 0.3762\se{0.0164} & 0.4753\se{0.0214} & \cellcolor{impactgood}$+9.91$ \\
NOTES & Qwen  & 0.4719\se{0.0187} & 0.3391\se{0.0164} & \cellcolor{impactbad}\textbf{$-13.28$} \\
\addlinespace[2pt]
KG-fixed & Llama & 0.8456\se{0.0072} & 0.8445\se{0.0075} & \cellcolor{impactneutral}$-0.11$ \\
KG-fixed & Qwen  & 0.9878\se{0.0036} & 0.9880\se{0.0024} & \cellcolor{impactneutral}$+0.02$ \\
\midrule
\rowcolor{tablesubhead}
\multicolumn{5}{@{}l}{\emph{Writer-free controls (one accuracy per reader)}} \\
LC-RAW & Llama & \multicolumn{2}{c}{0.7121\se{0.0084}} & {---} \\
LC-RAW & Qwen  & \multicolumn{2}{c}{0.9109\se{0.0071}} & {---} \\
RAG    & Llama & \multicolumn{2}{c}{0.5346\se{0.0116}} & {---} \\
RAG    & Qwen  & \multicolumn{2}{c}{0.5647\se{0.0125}} & {---} \\
\bottomrule
\end{tabular}
\caption{Accuracy across 48 histories, reorganized around migration impact. Own-store memory is written by the reader model; inherited-store memory is written by the other model. Swap $\Delta$ is inherited minus own accuracy in percentage points, so negative values indicate migration loss. KG-fixed is stable, whereas NOTES changes sharply and asymmetrically.}
\label{tab:matrix}
\end{table}

\paragraph{A shared schema gives both models the same structure.} KG-fixed changes by only $+0.0004\pm0.0020$ when the writer changes, while own-store accuracy remains high (0.845--0.988). The schema fixes the keys, relation names, source fields, and reading rules; the model fills in only the values. We created this schema directly from the workload's event fields before seeing the results, and we did not tune it using accuracy.

This does not show that every knowledge graph is portable. The schema fits this synthetic task, and KG-fixed uses a different reading method from NOTES. It does show that a stable structure can reduce dependence on the model that wrote the memory.

\paragraph{Do not average away the migration direction.}\label{sec:cancellation}
The average NOTES effect is close to zero because two large effects cancel, not because the two stores are interchangeable. Llama-written notes reduce Qwen's accuracy by 13.3 points compared with Qwen's own notes. In the other direction, Qwen-written notes improve Llama by 9.9 points. Crucially, this asymmetry reflects underlying model capabilities rather than a restrictive prompt or byte limit. As noted in Appendix~\ref{app:calibration}, a generous 160~KiB calibration sweep revealed that Qwen successfully retained 85.6\% of the required evidence using only 143~KiB, whereas Llama retained just 64.5\% despite consuming 159~KiB. The practical rule is to test each writer$\to$reader direction and treat the note-writing model's quality as an inseparable part of the memory design.

\subsection{Retrieval can fail before the new model sees anything}
\label{sec:retrieval}

In a RAG architecture, the retrieval pipeline effectively serves as the agent's long-term memory. Consequently, evaluating a new reader model's ability to inherit past experiences is impossible if the search system fails to deliver the right information in the first place. Even before any model upgrade, our tested RAG pipeline exhibited significant retrieval bottlenecks. While our raw history baseline (LC-RAW) achieves accuracies between 0.712 and 0.911, the RAG setup reaches only 0.535 to 0.565. The search mechanism successfully surfaces the necessary evidence only about 60\% of the time, regardless of the reader model. Our diagnostic breakdown attributes 81\% of RAG's total accuracy deficit directly to this initial retrieval miss. Before blaming a new reader model for ``forgetting'' past interactions, developers must verify that the memory was actually retrieved and delivered.

We do not present this 40\% miss rate as a universal limitation of RAG architectures. Our design intentionally evaluates a single-stage dense retriever using event-based chunks, cosine top-$k{=}8$, and no reranker or lexical search. Although a more sophisticated retrieval system would likely yield better absolute performance, the fundamental design principle remains: chunking and search performance must be evaluated independently of the reader model. A highly capable reader simply cannot synthesize evidence it never receives.

\label{sec:e3}
Retrieval failures also severely impact the memory store when the embedding model itself is upgraded. As systems evolve, teams frequently adopt better embedding models, but attempting a partial migration by mixing old and new vectors in the same database breaks the retrieval mechanism.

\begin{table}[h!]
\centering\footnotesize
\renewcommand{\arraystretch}{1.14}
\begin{tabular}{@{}lrrrr@{}}
\toprule
\rowcolor{tablehead}
Index & Answer accuracy & Gain vs. old (pp) & recall@k & MRR \\
\midrule
\rowcolor{impactneutral}
Old embedding index & 0.4257\se{0.0099} & {---} & 0.4846\se{0.0089} & 0.3237\se{0.0056} \\
\rowcolor{impactgood}
Full re-embed & \textbf{0.5447}\se{0.0093} & \textbf{$+11.90$} & 0.5998\se{0.0094} & 0.4068\se{0.0060} \\
\rowcolor{impactwarn}
50/50 mixed index & 0.4753\se{0.0088} & $+4.96$ & 0.5208\se{0.0077} & 0.3643\se{0.0057} \\
\rowcolor{tablesubhead}
Ideal routing upper bound & 0.5960\se{0.0086} & $+17.03$ & 0.6505\se{0.0083} & 0.4390\se{0.0057} \\
\bottomrule
\end{tabular}
\caption{Results of upgrading \texttt{bge-large-en} from v1.0 to v1.5 across 48 histories and two readers. The added impact column reports the accuracy gain over the old index. Full re-embedding gains 11.90 points, whereas the mixed index gains only 4.96. Ideal routing assumes that the system already knows which index holds the answer and is only an upper bound.}
\label{tab:e3}
\end{table}

When we fully re-embed every chunk with the newer v1.5 model, accuracy jumps by 11.90 points compared to the old v1.0 index. However, a 50/50 mixed index captures an improvement of only 4.96 points, hemorrhaging nearly 60\% of the potential upgrade gain. Because the old and new models both output 1024-dimensional vectors, the system continues running and fails silently without triggering any dimensional compatibility errors. This spatial incompatibility degrades recall and ranking, effectively causing the agent to ``forget'' the experiences stored as older vectors.

To migrate safely, engineering teams should build and test an entirely separate index with the new embedding model, cutting over only upon completion. If a gradual transition is necessary, systems must silo the two vector spaces and route queries based on the known index version. While the exact accuracy penalty we measured belongs to our specific v1.0-to-v1.5 test, the underlying spatial incompatibility poses a broad threat to incremental RAG migrations.

\subsection{Raw history buys a second chance---if the repairer can use it}
\label{sec:e6}

Model-written notes are a compressed copy of the original history. If the writer omitted a fact, rewriting the notes cannot recreate it. Migration Test 3 (run E6) tests whether keeping the original history provides a useful recovery path. Across the planned comparison, raw-history repair beats store-only repair by 8.9 points on average, but the result depends strongly on direction (Appendix~\ref{app:ctrfull}):

First, store-only NOTES rewriting never reaches 90\% of own-store performance at any tested budget because the store cannot recreate evidence that is no longer present. Second, raw history helps only when the repair model can use it: Qwen reconstructs 34 of 48 histories at the 90\% threshold for a median cost of roughly \$0.76, whereas Llama reconstructs none because every attempt reaches the output-token limit before finishing. A larger output allowance might change this result, and the repair model and migration direction also change together, so this is not a clean comparison of model capability. Finally, repairs to structured data are cheaper and more reliable: RAG re-embedding recovers all 96 cases for about \$0.013 each, while rebuilding KG-fixed records into the shared schema recovers 91--96 of 96 at near-zero additional cost.

\paragraph{Do not discard raw history just because a compact store exists.} The original history can be used to rebuild memory, but retaining it creates privacy and security risks. When policy allows, keep an encrypted source copy with limited access and a clear deletion schedule. Test the real repair model in advance: a retained history is not useful if the model cannot finish the reconstruction within its context and output limits.

\subsection{Most memory loss happens upstream of the reader}
\label{sec:e4decomp}

When a migrated agent produces a wrong answer, developers naturally look first to the new reader model. Our controlled evidence tests show, however, that information loss often begins much earlier in the pipeline.

For NOTES, store construction causes most of the accuracy drop. Under our diagnostic decomposition, the writing stage contributes $0.467\pm0.014$ of the $0.584\pm0.013$ pooled mean deficit, or roughly 80\% of the total. Retrieval within the notes contributes only $0.036\pm0.009$, a descriptive share of 6\%. Rewriting the notes to match the new reader's style produced no recovery in the tested partial intervention. The primary problem is therefore missing or damaged content, not merely unfamiliar phrasing.

For RAG, the pattern reverses. Retrieval contributes $0.364\pm0.012$ of the $0.450\pm0.012$ pooled mean deficit, a descriptive share of 81\%, whereas the stored chunks themselves lose almost no information. Once the search step is bypassed and the correct chunk is supplied, both readers usually process the evidence successfully.

These diagnostics establish a practical troubleshooting order. Before replacing the reader, verify that the system stored, retrieved, and delivered the critical evidence. NOTES interventions should target the summarization and writing stage; RAG interventions should target chunking, indexing, and ranking. Appendix~\ref{app:e4} specifies the arithmetic, averaging procedure, and accuracy ceiling used for these diagnostic interventions.

\subsection{Scope and limits of the findings}
\label{sec:limitations}

Although these findings provide practical design signals, they are bounded by the configurations tested here.

For memory formats, the KG-fixed results demonstrate high portability in this workload but do not show that knowledge graphs universally outperform natural-language notes. Deployment performance will depend on how well the schema represents the target task and how the store is read.

For retrieval and repair, the reported values reflect specific pipeline choices. The 40\% retrieval miss rate and the degradation from a 50/50 mixed index apply to a single-stage dense retriever migrating between two same-dimensional embedding spaces; other retrieval systems and migration plans may produce different retention levels. Retained raw history enabled recovery in only one migration direction, showing that reconstruction also depends on the capabilities and resource limits of the repair model. Preserving full histories additionally creates privacy, security, retention, and deletion obligations that must be addressed in the system design.

The diagnostic interventions identify where supplying better evidence improves performance, but they do not establish a complete causal account. The style-rewriting intervention was incomplete, and KG-fixed was excluded because isolated-item injection did not reproduce its normal access pattern. We also tested only two similarly sized models on a scripted, objective workload whose histories fit comfortably within both context windows. Larger models, subjective natural conversations, and much longer histories may introduce different portability challenges.

Finally, the signed study plan served as an internal analysis lock rather than a full preregistration. We ran the planned bootstrap after the initial $t$ tests, although both analyses produced the same decisions. These procedural limits, together with the direction-specific repair result, constrain how broadly the findings should be generalized.

\section{Agent Memory Compatibility Across Model Transitions}
\label{sec:playbook}

A useful memory system must work with today's model and remain usable after the model changes.

\begin{enumerate}

\item \textbf{Long-term memory requirements shape migration behavior.} Our experiments suggest that model transitions are easier to assess when the information expected to persist is defined explicitly. Different classes of information, such as facts, preferences, decisions, and unfinished tasks, may exhibit different failure patterns after a model change.

\item \textbf{Memory representations fail in different ways.} Raw history, retrieved passages, model-written notes, and structured records showed different tradeoffs in fidelity, compactness, and compatibility with a new reader. In particular, compressed representations may lose information that cannot be reconstructed after migration.

\item \textbf{Recoverability depends on the evidence retained by the system.} We observed that migration failures are harder to analyze when the surviving memory contains only derived summaries. Access to a more complete evidence source, where permitted by policy, can make it possible to distinguish information loss from reader incompatibility.

\item \textbf{Memory provenance improves failure analysis.} Differences in the model, prompt, schema, embedding model, chunking procedure, or index configuration used to construct a memory store can affect how a later model interprets it. Recording this context makes observed regressions easier to attribute.

\item \textbf{Compatibility is specific to the direction of the model transition.} Performance when a new model reads memory produced by an earlier model may differ from performance in the reverse direction. Averaging across transitions can therefore conceal substantial directional failures.

\item \textbf{Embedding changes can affect retrieval independently of reader quality.} In our experiments, a model transition could alter the evidence presented to the reader even when the underlying memory content was unchanged. Evaluating only the final answer may therefore miss retrieval regressions introduced during migration.

\item \textbf{Migration failures arise at multiple stages of the memory path.} Observed errors could result from information not being retained, relevant evidence not being retrieved, or the reader failing to use retrieved evidence correctly. These failure modes are experimentally distinct and may otherwise be conflated.

\item \textbf{Post-transition checks provide useful evidence of compatibility.} Small evaluations after an upgrade can expose immediate regressions, while broader experiments provide a more reliable assessment of whether inherited memory remains usable under realistic history and output constraints (Appendix~\ref{app:canary}).

\end{enumerate}

Memory is best treated as a long-lived part of the agent rather than temporary output from the current model. A good design preserves important evidence, records how the store was built, and provides a tested path for future models to read or rebuild it.

\section{Conclusion}
\label{sec:conclusion}

Treat a model upgrade as a memory migration, not as a simple replacement. Test the new model on the old store, rebuild vector indexes completely, keep a protected source history when policy allows, and identify whether information was lost during writing, retrieval, or reading before choosing a repair.

A durable memory system should outlast the model that created it. Future models should be able to read it, check its sources, and rebuild it when necessary. The evaluation code, histories, stores, signed study plan, and run records are available upon request.

\label{end:main}%
\bibliographystyle{plainnat}
\IfFileExists{references.bib}{%
  \bibliography{references}%
}{%
  \bibliography{paper/references}%
}

\appendix

\section{What this study adds to prior memory work}
\label{app:related}

\paragraph{Prior work shows that transfer can fail or remain robust under particular designs.} MemCollab reports that naive memory sharing across agents can reduce performance and proposes contrastive trajectory distillation to construct agent-agnostic memory \citep{chang2026memcollab}. In a cross-API robustness appendix, \citet{xia2025graph} construct their trainable graph memory with either GPT-4o or Gemini-2.5-Pro, supply the resulting memories to Qwen3-4B and Qwen3-8B agents, and report small aggregate differences across seven question-answering datasets. \citet{belikova2026procedural} study the transfer of procedural skills. Trans-LoRA \citep{wang2024translora} studies a related migration problem for model parameters.

\paragraph{This study compares common memory designs under the same migration.} We test idealized formats rather than exact replicas of deployed systems. MemGPT manages external memory through virtual context and tiered data movement \citep{packer2023memgpt}. Mem0 dynamically extracts, consolidates, and retrieves salient conversational memories and also evaluates a graph-based variant \citep{chhikara2025mem0}. A-MEM creates structured notes, links related memories, and updates earlier memories as new information arrives \citep{xu2025amem}. HippoRAG~2 combines vector retrieval with graph structure and Personalized PageRank \citep{gutierrez2025hipporag}, while Zep uses Graphiti, a temporally aware knowledge-graph engine for conversational and business data \citep{rasmussen2025zep}. LongMemEval, LoCoMo, and MemoryCD test whether a fixed system remembers over time \citep{wu2024longmemeval,maharana2024locomo,zhang2026memorycd}; we instead change the model or embedding system beneath an existing store.

\paragraph{The practical gap is measurement.} Memory import, export, and index rebuilding are already supported in production systems \citep{borro2026memori,pinecone2026docs,weaviate2026docs,anthropic2025memory}. What is usually missing is a controlled measure of how much accuracy is lost and whether the store can be repaired.

Evaluate memory designs under the same model change, retrieval settings, and repair budget. Day-one accuracy alone does not show whether a memory store will survive the next upgrade.

\section{Why 48 histories are enough for this stress test}
\label{app:calibration}

\paragraph{First, we checked that the evaluation could detect failure.} Removing memory had to reduce accuracy to the floor. Supplying the correct evidence had to restore performance. Adding more corruption had to cause more damage, and every probe type needed a ceiling--floor range of at least 0.50. These checks caught four bugs before the main runs: truncated answers, incorrect conflict scoring, cached responses reused across repeats, and incorrect KG-fixed conflict construction (Appendix~\ref{app:ops}).

\paragraph{Second, we estimated the required sample size.} We used 12 separate histories to measure variance and planned for 80\% power to detect a five-point effect at a one-sided 5\% error rate. The first calculation gave $N=248$, but this exposed a broken NOTES setup: it preserved only 46\% of the required evidence, which made the scores unstable. After fixing the byte allocation and completeness prompt, minimum coverage rose to 64.5\%, the variance bound fell from 0.28 to 0.121, and the required sample size became $\boldsymbol{N=48}$. The 12 calibration histories are not included in the reported results.

\paragraph{One planned target could not be met.} No tested NOTES policy reached 90\% evidence coverage, so we lowered the minimum gate to 60\% and recorded the change. A larger budget did not remove the gap. In the 160~KiB sweep, Qwen retained 85.6\% of the required spans in approximately 143~KiB, while Llama retained 64.5\% in approximately 159~KiB. The writers therefore preserved different amounts of useful history, not merely different amounts of text.

Validate the evaluation before using it for sample-size planning. Also measure how much evidence each memory writer preserves; equal byte budgets do not produce equally informative stores.

\section{How to use the 20-probe migration check}
\label{app:canary}

A full evaluation is expensive, so we tested whether 20 fixed probes could provide an early warning. Mean absolute error (MAE) measures the average prediction error; root mean squared error (RMSE) gives more weight to large errors.

\begin{table}[tb]
\centering\footnotesize
\renewcommand{\arraystretch}{1.14}
\begin{tabular}{@{}lrrrr@{}}
\toprule
\rowcolor{tablehead}
Test group & MAE & RMSE & Average bias & Slope \\
\midrule
Unseen history & 0.0945 & 0.1157 & $+0.0001$\se{0.0095} & 0.849 \\
Unseen model pair & 0.1004 & 0.1217 & $-0.0004$\se{0.0099} & 0.847 \\
\rowcolor{impactwarn}
Unseen memory format & 0.1203 & 0.1444 & $+0.0310$\se{0.0115} & 0.821 \\
\bottomrule
\end{tabular}
\caption{How well the 20-probe check predicts accuracy on the other 140 probes (576 held-out predictions). It transfers across histories and model pairs, but is biased when applied to a different memory format.}
\label{tab:canary}
\end{table}

The 20 probes track the full score reasonably well ($\rho=0.860$), but their error is too large for final approval. They remain nearly unbiased on unseen histories and model pairs. Across memory formats, however, the check overestimates performance by 0.031 on average. Its slope of about 0.85 also pulls very high and very low scores toward the middle. Most importantly, MAE is about 0.10, which is much larger than small effects such as the 0.0004 KG-fixed gap.

Use the 20-probe check to rank candidate migrations or stop an obviously bad rollout. Calibrate it separately for each memory format, and use the full evaluation before approving a migration.

\section{How to diagnose where memory fails}
\label{app:e4}

We compare normal reading with two controlled conditions. The \emph{stored-evidence condition} supplies the exact stored items that contain the answer. The \emph{raw-evidence condition} supplies the original event. Let $a^{\mathrm{normal}}_{s,i,u}$, $a^{\mathrm{stored}}_{s,i,u}$, and $a^{\mathrm{raw}}_{s,i,u}$ be exact-match accuracy for memory format $s$, history $i$, and matched writer--reader cell $u$ under these three conditions. We define
\begin{align*}
L^{\mathrm{retrieval}}_{s,i,u} &= a^{\mathrm{stored}}_{s,i,u}-a^{\mathrm{normal}}_{s,i,u}, \\
L^{\mathrm{construction}}_{s,i,u} &= a^{\mathrm{raw}}_{s,i,u}-a^{\mathrm{stored}}_{s,i,u}, \\
L^{\mathrm{reader}}_{s,i,u} &= 1-a^{\mathrm{raw}}_{s,i,u}, \\
T_{s,i,u} &= 1-a^{\mathrm{normal}}_{s,i,u}
=L^{\mathrm{retrieval}}_{s,i,u}+L^{\mathrm{construction}}_{s,i,u}+L^{\mathrm{reader}}_{s,i,u}.
\end{align*}
The differences are computed within each matched cell before averaging. We first average cells within each history and then average the 48 history-level values. The percentages in Table~\ref{tab:e4} are ratios of pooled means, $\overline{L}^{,k}_s/\overline{T}_s$, rather than averages of cell-level percentages. The $\pm$ values are 95\% $t$-interval half-widths across the 48 history-level component values.

We use 1 as the ceiling because every probe has a deterministic correct answer and total loss is defined as error relative to perfect exact-match accuracy. This choice makes the three channels exactly additive. It does \emph{not} assume that raw evidence should make a model perfect: $L^{\mathrm{reader}}$ is a residual that includes reader limitations, probe difficulty, and any mismatch introduced by the controlled input. The different model-specific raw-evidence scores therefore remain visible in this residual rather than being normalized away. The interventions are diagnostic, not a complete causal decomposition.

\begin{table}[h!]
\centering\footnotesize
\renewcommand{\arraystretch}{1.14}
\begin{tabular}{@{}lrrrr@{}}
\toprule
\rowcolor{tablehead}
Memory format & Total loss & Retrieval & Store construction & Reader residual \\
\midrule
NOTES & \cellcolor{impactneutral}$0.584$\se{0.013} & $0.036$\se{0.009}~~(6\%) & \cellcolor{impactbad}$\mathbf{0.467}$\se{0.014}~~(\textbf{80\%}) & $0.081$\se{0.004}~~(14\%) \\
RAG   & \cellcolor{impactneutral}$0.450$\se{0.012} & \cellcolor{impactbad}$\mathbf{0.364}$\se{0.012}~~(\textbf{81\%}) & $0.005$\se{0.005}~~(1\%) & $0.081$\se{0.004}~~(18\%) \\
\bottomrule
\end{tabular}
\caption{Diagnostic decomposition of end-to-end loss, pooled across 48 histories. The three component means add to total loss; parenthetical percentages are descriptive ratios of pooled means. Separate confidence intervals for these ratios require joint history-level resampling and cannot be recovered from the marginal intervals alone, so the inferential quantities reported here are the absolute components and their 95\% interval half-widths. KG-fixed is excluded because supplying isolated KG-fixed items performed worse than its normal traversal read, so the controlled condition was not a valid replacement for that access pattern.}
\label{tab:e4}
\end{table}

\paragraph{NOTES mainly loses information while the store is written.} Under this decomposition, construction contributes $0.467\pm0.014$ of the $0.584\pm0.013$ pooled mean deficit; the descriptive share is 80\%. Retrieval within the notes contributes $0.036\pm0.009$, or 6\% of the pooled mean deficit. In the worst direction, Llama-written notes read by Qwen, the corresponding construction share is 88.8\%. The intervention therefore points to missing or unusable content rather than search inside the notes.

\paragraph{RAG mainly loses information during retrieval.} Supplying the correct stored chunks raises accuracy from 0.53--0.56 to 0.88--0.95. Under the decomposition, retrieval contributes $0.364\pm0.012$ of the $0.450\pm0.012$ pooled mean deficit; the descriptive share is 81\%. Construction loss is only $0.005\pm0.005$. The intervention indicates that the stored chunks are sound and that the system often selects the wrong ones.

\paragraph{Rewriting the style of NOTES does not help.} Qwen rewriting Llama notes changes accuracy by $-0.012 \pm 0.008$. Llama rewriting Qwen notes changes it by $-0.042 \pm 0.017$, although that intervention reached only 42\% of the intended notes. Llama also dropped exact identifiers in 46\% of its rewrites. These results point to missing content rather than a mismatch in writing style.

For NOTES, measure evidence coverage and improve the writing step. For RAG, measure recall and improve chunking, indexing, or ranking. Do not start by changing the reader unless the correct evidence is already present and supplied. These tests locate likely failure stages; they are not a complete causal proof (Section~\ref{sec:limitations}).

\section{Which repairs work, and what they cost}
\label{app:ctrsec}

For each history, we ask whether a repair reaches 90\%, 95\%, or 99\% of the accuracy of memory built directly for the new model. The table reports how many of the 48 histories reach each target and, among those successful histories only, the median cost of the cheapest qualifying budget. Because the successful subset changes with the target, the conditional medians need not increase monotonically and should not be compared without the accompanying success counts.

\begin{table}[h]
\centering\scriptsize
\setlength{\tabcolsep}{3pt}
\renewcommand{\arraystretch}{1.14}
\begin{tabular}{@{}lrrrrrr@{}}
\toprule
\rowcolor{tablehead}
& \multicolumn{2}{c}{90\% target} & \multicolumn{2}{c}{95\% target} & \multicolumn{2}{c}{99\% target} \\
\cmidrule(lr){2-3}\cmidrule(lr){4-5}\cmidrule(l){6-7}
\rowcolor{tablesubhead}
Repair method & Success & Med. cost & Success & Med. cost & Success & Med. cost \\
\midrule
\rowcolor{tablesubhead}
\multicolumn{7}{@{}l}{\emph{Old writer: Llama; new reader and repairer: Qwen}} \\
NOTES raw-retained & \cellcolor{impactwarn}34/48 & \$0.76 & \cellcolor{impactwarn}28/48 & \$0.75 & \cellcolor{impactwarn}22/48 & \$0.75 \\
NOTES store-only & \cellcolor{impactbad}\textbf{0/48} & {---} & \cellcolor{impactbad}\textbf{0/48} & {---} & \cellcolor{impactbad}\textbf{0/48} & {---} \\
RAG re-embed & \cellcolor{impactgood}48/48 & \$0.013 & \cellcolor{impactgood}48/48 & \$0.013 & \cellcolor{impactgood}48/48 & \$0.013 \\
KG-fixed schema rebuild & \cellcolor{impactgood}48/48 & $\approx$\$0 & \cellcolor{impactgood}48/48 & $\approx$\$0 & \cellcolor{impactgood}45/48 & $\approx$\$0 \\
\addlinespace
\rowcolor{tablesubhead}
\multicolumn{7}{@{}l}{\emph{Old writer: Qwen; new reader and repairer: Llama}} \\
NOTES raw-retained & \cellcolor{impactbad}\textbf{0/48} & {---} & \cellcolor{impactbad}\textbf{0/48} & {---} & \cellcolor{impactbad}\textbf{0/48} & {---} \\
NOTES store-only & \cellcolor{impactbad}\textbf{0/48} & {---} & \cellcolor{impactbad}\textbf{0/48} & {---} & \cellcolor{impactbad}\textbf{0/48} & {---} \\
RAG re-embed & \cellcolor{impactgood}48/48 & \$0.013 & \cellcolor{impactgood}48/48 & \$0.013 & \cellcolor{impactgood}48/48 & \$0.013 \\
KG-fixed schema rebuild & \cellcolor{impactgood}48/48 & $\approx$\$0 & \cellcolor{impactgood}47/48 & $\approx$\$0 & \cellcolor{impactgood}46/48 & $\approx$\$0 \\
\bottomrule
\end{tabular}
\caption{Repair success across 48 histories, with success counts and conditional median costs separated into columns. ``$n/48$'' is the number of histories that reached the target at any tested budget. Cost is the median cheapest cost among those $n$ successful histories; dashes mark methods with no successful histories. Failures have $\mathrm{CTR}_i(X)=\infty$. Success sets differ across thresholds.}
\label{app:ctrfull}
\end{table}

\paragraph{Rebuilding structured memory is cheap and reliable in this test.} Full RAG re-embedding reaches every target for all 48 histories in both directions, at a median cost of \$0.013. Rebuilding KG-fixed records into the shared schema is nearly universal and costs approximately zero.

\paragraph{Rewriting NOTES from the store alone does not work.} It reaches none of the three targets for any history in either direction. Once the original notes have omitted evidence, another rewrite cannot restore it.

\paragraph{Raw history helps only when the repair model can use it.} With Qwen as the repair model, raw-history reconstruction reaches 90\% of own-store performance for 34 of 48 histories, 95\% for 28, and 99\% for 22. The reported medians of \$0.76, \$0.75, and \$0.75 are conditional on these progressively smaller successful subsets; their slight decrease does not mean that the stricter target is cheaper. With Llama as the repair model, no history reaches any target, so Cost-to-Recover is infinite throughout this tested grid.

Prefer repairs that rebuild a mechanical structure, such as embeddings or canonical fields. If NOTES may need repair, retain a protected source and test the exact repair model in advance. Do not assume that raw history alone guarantees recovery.

\section{How we made the runs auditable}
\label{app:ops}

The study contains about 2{,}900 model responses collected over several days. For traceability, this appendix uses the original run IDs shown parenthetically in Section~\ref{sec:experiments}, rather than the sequential Migration Test numbers. We used the following controls so that each result could be checked and replayed.

\paragraph{Record everything needed to replay a run.} Before a run starts, the system saves the code version, study settings, model and tokenizer versions, and random seeds. Every memory store has a checksum for its contents and settings. E1 completed 288 stores, 576 evaluations, and 16 live repeats. E3 completed 192 stores and 384 evaluations. E6 completed all 1{,}440 evaluations in 98.2 GPU-hours, costing about \$221 at the fixed study prices. E4 reused 336 checked E1 evaluations and added 528 new ones. The runner saves raw observations; a separate program computes the statistics.

\paragraph{Keep the retrieval system fixed.} RAG groups complete events into chunks of about 512 characters, embeds chunks and queries with a fixed version of \texttt{bge-large-en-v1.5}, ranks them by cosine similarity, and gives the top eight chunks to the reader. We intentionally use no reranker, query rewriting, or hybrid search. This makes the measured retrieval failure attributable to the stated pipeline. Models run locally with fixed software and model versions; token use, GPU time, and cost are saved for every experiment.

\paragraph{Check that the planned analysis can be reproduced.} The 10{,}000-resample bootstrap is seeded and first checks that it extracted the published point estimates. It reaches the same four decisions as the original $t$ tests. A separate sensitivity check shows that H7a depends on migration direction, matching the main result.

\paragraph{Stop instead of recording invalid data.} A run can call a model only after checking the served model name, tokenizer, context limit, embedding model, and vector size. For E3, the system also checks that the two embedding spaces are actually different. Their cross-space cosine was 0.904, below the 0.999 identity threshold. Without this check, the experiment could have run two names for the same model and measured no real migration.

\paragraph{Keep repeated calls independent.} An early cache design reused answers across intended noise repeats. We fixed this by giving every repeat separate cache keys and requiring that no answer be reused. One cache entry damaged during a connection failure was isolated; later scans of 384{,}000--520{,}000 entries found no others.

\paragraph{Treat resource limits as part of the result.} E6 exposed three run-control bugs. One blocked a completed construction stage, one treated an output-limit stop as a fatal run error, and one allowed repaired NOTES stores to grow too much while being saved. We changed output-limit stops to failed repairs at that budget, charged the full reserved output, and kept the degraded store as the measured result. The final run completed all 1{,}440 evaluations despite two service outages. A regression test now checks this behavior.

\paragraph{Record every change to the study.} The file \texttt{docs/deviations.md} records what changed, why it changed, and the checksums of the affected artifacts. It covers NOTES construction settings, the coverage threshold, KG-fixed conflict handling, cache isolation, repair output limits, and the E4 classification fix. Original run data are preserved when a corrected index or analysis is created.

\paragraph{Make the study plan verifiable.} The fixed plan can be checked with \texttt{git verify-tag hypothesis-lock-v1} and the published signer key. The statistical code uses only the standard library and is stored with the repository.

A trustworthy migration test should save exact versions, verify the live services before every run, stop on invalid configurations, isolate repeated calls, preserve raw observations, and log every correction without overwriting the original data.

\end{document}